\documentclass[runningheads]{llncs}
\usepackage{amsmath,graphicx,xcolor,amsfonts}
\usepackage[inline]{enumitem}
\usepackage{makeidx}  
\usepackage{amssymb}
\usepackage{subcaption}
\usepackage{siunitx}
\usepackage{url}
\usepackage{tabularx}

\usepackage{etoolbox}
\makeatletter
\newlength{\parboxtodim}
\patchcmd{\@iiiparbox}
{\hsize}
{\ifx\relax#2\else\setlength{\parboxtodim}{#2}\fi\hsize}
{}{}
\makeatother

\def\mathunderline#1#2{\color{#1}\underline{{\color{black}#2}}\color{black}}
\newcolumntype{C}[1]{>{\centering\let\newline\\\arraybackslash\hspace{0pt}}m{#1}}

\begin{document}

\title{Deep Neural Dynamic Bayesian Networks applied to EEG sleep spindles modeling\thanks{Supported by NSF CAREER grant 1453756. All code and data can be accessed at https://github.com/carlosloza/DNDBN}}
\titlerunning{Deep Neural Dynamic Bayesian Networks applied to EEG sleep spindles}
%
\author{Carlos A. Loza\orcidID{0000-0002-0509-5104} \and
	Laura L. Colgin\orcidID{0000-0002-4853-8913}}
\authorrunning{C. Loza and L. Colgin}
%

\institute{Department of Neuroscience, Center for Learning and Memory, \\The University of Texas at Austin\\
	\email{carlos.loza@utexas.edu, colgin@mail.clm.utexas.edu}}
%
%
%
%
\maketitle
\begin{abstract}
We propose a generative model for single--channel EEG that incorporates the constraints experts actively enforce during visual scoring. The framework takes the form of a dynamic Bayesian network with depth in both the latent variables and the observation likelihoods---while the hidden variables control the durations, state transitions, and robustness, the observation architectures parameterize Normal--Gamma distributions. The resulting model allows for time series segmentation into local, reoccurring dynamical regimes by exploiting probabilistic models and deep learning. Unlike typical detectors, our model takes the raw data (up to resampling) without pre--processing (e.g., filtering, windowing, thresholding) or post--processing (e.g., event merging). This not only makes the model appealing to real--time applications, but it also yields interpretable hyperparameters that are analogous to known clinical criteria. We derive algorithms for exact, tractable inference as a special case of Generalized Expectation Maximization via dynamic programming and backpropagation. We validate the model on three public datasets and provide support that more complex models are able to surpass state--of--the--art detectors while being transparent, auditable, and generalizable.
\end{abstract}
%
%
\section{Introduction}
\label{sec:intro}

Sleep spindles are a hallmark of stage 2 non--REM sleep. They are the result of interactions between GABAergic reticular neurons and excitatory thalamic cells \cite{steriade1993thalamocortical}. Their proposed functions include memory consolidation \cite{schabus2004sleep}, cortical development, and potential biomarkers for psychiatric disorders \cite{ferrarelli2007reduced}. Hence, proper detection and modeling using Electroencephalogram (EEG) are crucial.

A sleep spindle is defined as an oscillatory burst in the range 11--15 Hz (sigma band) with duration between 0.5 and 2 s., waxing--waning envelope, and maximal in amplitude in central EEG electrodes. EEGers usually adhere to clinical manuals \cite{rechtschaffen1968manual,iber2007aasm,niedermeyer2005electroencephalography} to visually categorize multi--channel EEG traces into sleep stages. They are also trained to identify non--brain--related artifacts (e.g., eye and muscle activity). Currently, the large amount of data calls for machine learning techniques to guide principled, human--like automatic EEG scorers.

Most automatic sleep spindles detectors comprise four stages: pre--processing, decomposition, decision making, and feature extraction \cite{coppieters2016sleep}. The first stage usually involves bandpassing and artifact rejection, while the second stage applies wavelets or other windowing techniques for quasi--stationary processes. Decision making takes the form of hard--thresholding plus cross--validation, and lastly, feature extraction characterizes the events for further analysis. A myriad of approaches use slight variations of this pipeline \cite{devuyst2011automatic,wendt2012validation,martin2013topography,parekh2014sleep} (reviews in  \cite{coppieters2016sleep}, \cite{warby2014sleep}). 

Even though this methodology provides adequate results, it suffers from three main drawbacks:  \begin{enumerate*}[label=\roman*)]
	\item analysis based on filtered EEG,
	\item lack of theoretical foundations for hyperparameter settings, and
	\item reported results are often in--sample predictions, which is prone to overfitting.
\end{enumerate*}
The first point not only differs from human--like scoring, it is also computationally expensive for online applications. The latter points are consequences of a model--less framework; that is, amplitude thresholds are data (and scorer) dependent, which severely limits generalization. 

Instead, we propose a generative model with the following components: \begin{enumerate*}[label=\roman*)]
	\item robustness against artifacts via observations modeled as conditional heteroscedastic non--linear generalized t likelihoods,
	\item high--capacity deep learning architectures that parameterize such likelihoods,
	\item reoccurring modes with distinctive dynamics that characterize non--spindle and spindle regimes,
	\item semi--Markovian states that generalize the geometric regime durations of Hidden Markov Models (HMM),
	\item tractable exact inference via message passing routines (unlike variational approximations \cite{johnson2016composing,ebbers2017hidden,krishnan2017structured,dong2020collapsed}), and
	\item uncertainty quantification via posterior probabilities.
\end{enumerate*}
The result is a model that combines probabilistic modeling with deep learning in an effort to mimic the constraints EEGers actively enforce when visually scoring EEG: a Deep Neural Dynamic Bayesian Network.

We validate the model on three datasets from the DREAMS database \cite{stephanie_devuyst_2005_2650142} and achieve comparable or better performance than the state--of--the--art. The results open the door to more complex models and sophisticated inference in the future (e.g., real--time detection). The paper continues as follows: Section 2 details the model, Section 3 presents the results, and Section 4 concludes the paper.

\section{Generative model for EEG}
\label{sec:methods}
\begin{figure}[tb]
	\centering
	\begin{subfigure}[t]{0.4793\linewidth}
		\includegraphics[width=1\columnwidth]{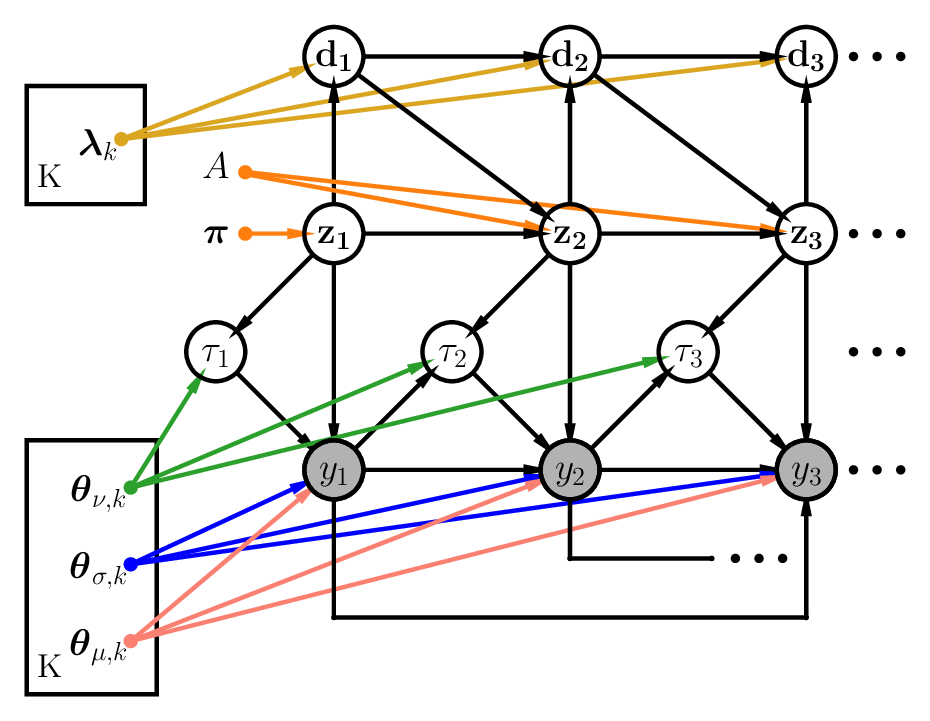}
		\label{fig:GraphicalModel}
	\end{subfigure}
	\begin{subfigure}[t]{0.4793\linewidth}
		\includegraphics[width=1\columnwidth]{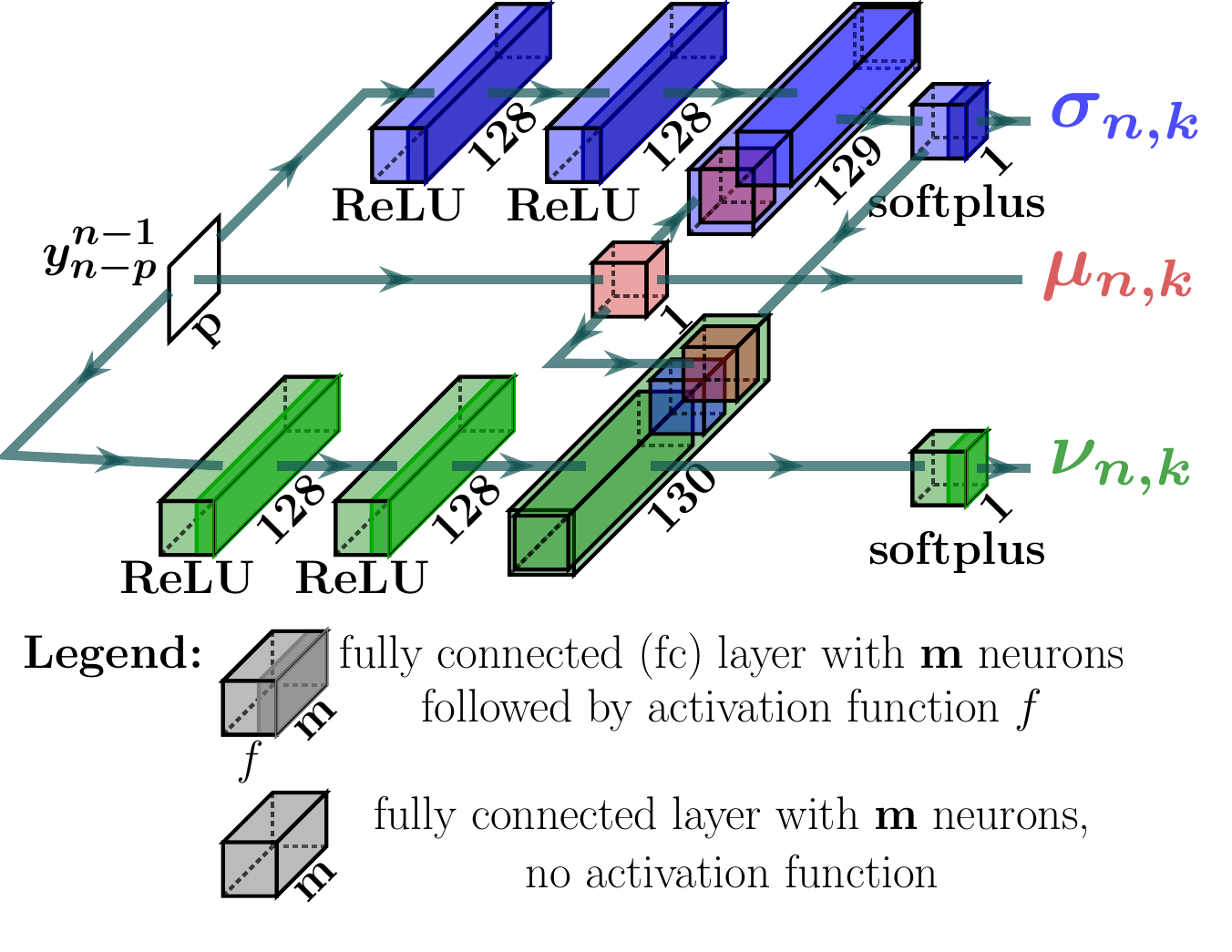}
		\label{fig:DeepObservationModels}
	\end{subfigure}
	\caption{\textbf{Left.} Generative model for EEG. Observations are shaded. \textbf{Right.} Deep networks for Normal--Gamma models (colors paired with graphical model).}
	\label{fig:DNDBN}
\end{figure}

Let $\mathbf{y} = \{y_n\}_{n=1}^N$ be an ordered collection of $N$ observed random variables over time. For our case, $\mathbf{y}$ is a single--channel, single--trial EEG trace embedded into a dynamical system with $K$ discrete states or labels (non--spindle and spindle for our case). States are represented by a $K$--dimensional multinomial $\mathbf{z_n}$. Additionally, the $D$--dimensional multinomial, $\mathbf{d_n}$, is a counter variable that keeps track of the remaining duration of the current mode. Given a regime $k$, we pose each $y_n$ as a Normal random variable with location, $\mu_{n,k}$, and scale, $\sigma_{n,k}$, parameterized by functions (either linear or non--linear) of the previous $p$ samples, $y_{n-p}^{n-1}$, where $p$ is the autoregressive order of the model. In particular, $\sigma_{n,k}$ is modulated by the conditionally Gamma distributed hidden variable, $\tau_n$, that accounts for robustness. Likewise, the parameter of the Gamma distribution, $\nu_{n,k}$, depends on $y_{n-p}^{n-1}$. In short, given a state $k$, $\tau_n$ is marginally Gamma distributed and $y_n$ is conditionally Normal given $\tau_n$; thus, $y_n$ is marginally generalized t distributed:
\begin{align}
	\tau_n | \mathbf{z_n}, y_{n-p}^{n-1}; \{\text{\boldmath$\theta$\unboldmath}_{\nu, k}\}_{k=1}^K &\sim \text{Gamma}\Big(\frac{\nu_{n,k}}{2},
	\frac{\nu_{n,k}}{2}\Big)\\
	y_n | \tau_n, \mathbf{z_n}, y_{n-p}^{n-1};  \{\text{\boldmath$\theta$\unboldmath}_{\mu,k}\}_{k=1}^K, \{\text{\boldmath$\theta$\unboldmath}_{\sigma,k}\}_{k=1}^K  &\sim \mathcal{N} \bigg( \mu_{n,k}, \frac{\sigma_{n,k}}{\sqrt{\tau_n}} \bigg)
	\label{eq:Obs}
\end{align}
where $\text{\boldmath$\theta$\unboldmath}_{\mu, k}$, $\text{\boldmath$\theta$\unboldmath}_{\sigma, k}$, and $\text{\boldmath$\theta$\unboldmath}_{\nu, k}$ parameterize a deep neural network with dedicated architectures, $F_{\text{\boldmath$\theta$\unboldmath}_{\mu, k}}(y_{n-p}^{n-1})$, $G_{\text{\boldmath$\theta$\unboldmath}_{\sigma, k}}(y_{n-p}^{n-1})$, $H_{\text{\boldmath$\theta$\unboldmath}_{\nu, k}}(y_{n-p}^{n-1})$, for $\mu_{n,k}$, $\sigma_{n,k}$, and $\nu_{n,k}$, respectively (Fig. \ref{fig:DNDBN}). The deep observation models allow for rich representations, whereas the deep 3--level hidden state accounts for robustness and regime durations beyond the usual geometric paradigm of HMM. Lastly, the autoregressive structure is key to capture local temporal dependencies (e.g.,  sleep spindles). The result is the Deep Neural Dynamic Bayesian Network (DNDBN) of Fig. \ref{fig:DNDBN}.

We use a 1-of-$K$ notation for $\mathbf{z_n}$ and $\mathbf{d_n}$ (e.g., $z_{n,k}$ means the $k$--th entry of $\mathbf{z_n}$ is 1). $\mathbf{z_1}$ is parameterized by \boldmath$\pi$\unboldmath, while $\{\mathbf{z_{n}}\}_{n=2}^N$ obey semi--Markovian dynamics:
\begin{equation}
p(\mathbf{z_n}|\mathbf{z_{n-1}},\mathbf{d_{n-1}}; A) =
\left\{
\begin{array}{ll}
\delta(\mathbf{z_n},\mathbf{z_{n-1}}) \qquad \, ,d_{n-1,1} > 1   \\
p(\mathbf{z_n}|\mathbf{z_{n-1}}; A) \quad \, ,d_{n-1,1} = 1\\
\end{array}
\right.
\label{eq:zState}
\end{equation}
where $A$ is a $K \times K$ stochastic matrix, and $\delta(\cdot,\cdot)$ is the Kronecker delta function. In words, the hidden label, $\mathbf{z_n}$, transitions to a different regime only when the counter variable is equal to 1; otherwise, it remains in the same mode. 

Similarly, $\mathbf{d_n}$ samples a new duration only after a regime ends; otherwise, it decreases by one (i.e., an explicit duration hidden semi--Markov model \cite{ferguson1980variable,yu2010hidden}).
\begin{equation}
p(\mathbf{d_n}|\mathbf{z_n},\mathbf{d_{n-1}}; \{\text{\boldmath$\lambda$\unboldmath}\}_{k=1}^K )=
\left\{
\begin{array}{ll}
\delta(d_{n,j-1},d_{n-1,j}) \qquad \qquad \qquad \quad \, ,d_{n-1,1} > 1   \\
p(\mathbf{d_n}|z_{n,k}; \{\text{\boldmath$\lambda$\unboldmath}\}_{k=1}^K) = p(\mathbf{d_n};\text{\boldmath$\lambda$\unboldmath}_k) \quad ,d_{n-1,1} = 1\\
\end{array}
\right.
\label{eq:dState}
\end{equation}
where $\{\text{\boldmath$\lambda$\unboldmath}\}_{k=1}^K $ is a collection of parameters for each regime durations.

Consequently, we can write the complete data log--likelihood over parameters, $\text{\boldmath$\theta$\unboldmath}=\{$\boldmath$\lambda$\unboldmath$, A,$\boldmath$\pi$\unboldmath$,\{\text{\boldmath$\theta$\unboldmath}_{\mu, k}\}_{k=1}^K,\{\text{\boldmath$\theta$\unboldmath}_{\sigma, k}\}_{k=1}^K,\{\text{\boldmath$\theta$\unboldmath}_{\nu, k}\}_{k=1}^K \}$, as (ignoring constants):

\begin{equation}
\begin{split}
&\log{p(\mathbf{y}, \text{\boldmath$\tau$\unboldmath}, \mathbf{z}, \mathbf{d}| \text{\boldmath$\theta$\unboldmath})} =
\sum_{k=1}^{K} \mathunderline{cyan}{z_{1,k}} \log\pi_k+ \sum_{i=1}^{D}\sum_{k=1}^{K} \mathunderline{yellow}{z_{1,k} d_{1,i}} \log p(\mathbf{d_1};\text{\boldmath$\lambda$\unboldmath}_k) +\\
&\sum_{n=2}^{N}\sum_{k=1}^{K}\sum_{j=1}^{K}  \mathunderline{green}{z_{n-1,j} z_{n,k} d_{n-1,1}} \log A_{j,k} + \sum_{n=2}^{N}\sum_{k=1}^{K} \mathunderline{green}{z_{n,k} d_{n-1,1}} \log p(\mathbf{d_n};\text{\boldmath$\lambda$\unboldmath}_k) +\\ 
&\sum_{n=1}^{N}\sum_{k=1}^{K} \mathunderline{cyan}{z_{n,k}}\bigg\{ 
\frac{\Gamma(\frac{\nu_{n,k} + 1}{2})}{\Gamma(\frac{\nu_{n,k}}{2})\sqrt{\pi\nu_{n,k}}\sigma_{n,k}}
\bigg(1 + \frac{1}{\nu_{n,k}}
\Big(\frac{y_n - \mu_{n,k}}
{\sigma_{n,k}}\Big)^2\bigg)^
{-\big(\frac{\nu_{n,k}+1}{2}\big)} \bigg\}
\end{split}
\label{eq:loglike}
\end{equation}
where $\Gamma(\cdot)$ is the gamma function, $A_{j,k}$ is the $j$--th row, $k$--th column entry of $A$, and $\mathbf{z}=\{\mathbf{z_n}\}_{n=1}^N$ (similar for $\mathbf{d}$ and \boldmath$\tau$\unboldmath). Three tasks are key: \begin{enumerate*}[label=\alph*)] 
	\item given $\text{\boldmath$\theta$\unboldmath}$, calculate the marginal log--likelihood (LL) of the observations (i.e., $p(\mathbf{y};\text{\boldmath$\theta$\unboldmath})$), 
	\item given $\mathbf{y}$ and $\text{\boldmath$\theta$\unboldmath}$, estimate the most likely hidden state sequence, and 
	\item given $\mathbf{y}$, estimate $\text{\boldmath$\theta$\unboldmath}$.
	\end{enumerate*} We refer to the tasks as LL calculation, inference, and learning, respectively. 

In terms of interpretability, the location networks, $\{F_{\text{\boldmath$\theta$\unboldmath}_{\mu, k}}(y_{n-p}^{n-1})\}_{k=1}^K$, are reduced to single fully--connected layers with no non--linearities. This design choice not only renders linear autoregressive models (i.e., infinite impulse response filters), but it also helps with validation while keeping the model more interpretable without sacrificing performance (as shown in the next sections).

\subsection{Learning model parameters}
Maximizing eq. \eqref{eq:loglike} analytically is intractable; rather, we exploit a variation of the Generalized Expectation--Maximization (GEM) algorithm for HMM \cite{rabiner1989tutorial}. We implement coordinate ascent on the log--likelihood: first, $\text{\boldmath$\theta$\unboldmath}$ is fixed and we obtain expectations of $\log{p(\mathbf{y}, \text{\boldmath$\tau$\unboldmath}, \mathbf{z}, \mathbf{d}| \text{\boldmath$\theta$\unboldmath})}$ under the posterior distribution of the latent variables given $\mathbf{y}$ (underlined variables in eq. \eqref{eq:loglike}). Then, $\text{\boldmath$\theta$\unboldmath}$ is optimized (either globally or locally) keeping the expectations fixed. This constitutes one GEM iteration and guarantees an increase of the marginal LL of the training data. GEM continues until convergence to local optima (tracked by successive LLs).

The factorizations induced by the model yield closed--form solutions of the expectations without appealing to variational approximations that usually require recognition networks to compute local potentials \cite{kingma2013auto,johnson2016composing,ebbers2017hidden,krishnan2017structured,dong2020collapsed}. Hence, our proposed learning routine does not suffer from the so--called amortization gap \cite{cremer2018inference} nor ``posterior collapse" \cite{he2019lagging,dieng2019avoiding}. Next, for simplicity, we outline the GEM algorithm for the case of a single training sequence; yet, the batch case is straightforward.

For the E--step, let $\alpha_{\mathbf{z_n}, \mathbf{d_n}} = p(y_1^n, \mathbf{z_n}, \mathbf{d_n})$ be the joint probability of the first $n$ observations and the $n$--th hidden variables, and similarly $\beta_{\mathbf{z_n}, \mathbf{d_n}} = p(y_{n+1}^N| \mathbf{z_n}, \mathbf{d_n})$. Next, by induction and d--separation \cite{koller2009probabilistic} (no parameters to avoid clutter):
\begin{align}
	\begin{split}\label{eq:alpha}
		\alpha_{\mathbf{z_n}, \mathbf{d_n}} &=  p(y_n|\mathbf{z_n}) \sum_{\mathbf{d_{n-1}}} p(\mathbf{d_n}|\mathbf{z_n},\mathbf{d_{n-1}}) 
		\sum_{\mathbf{z_{n-1}}}  p(\mathbf{z_n}|\mathbf{z_{n-1}},\mathbf{d_{n-1}}) \alpha_{\mathbf{z_{n-1}}, \mathbf{d_{n-1}}} 
	\end{split}\\
	\begin{split}\label{eq:beta}
		\beta_{\mathbf{z_n}, \mathbf{d_n}} &=  \sum_{\mathbf{z_{n+1}}} p(y_{n+1}|\mathbf{z_{n+1}})  p(\mathbf{z_{n+1}}| \mathbf{z_{n}}, \mathbf{d_{n}})  
		\sum_{ \mathbf{d_{n+1}}} p(\mathbf{d_{n+1}}|\mathbf{z_{n+1}}, \mathbf{d_{n}})  \beta_{\mathbf{z_{n+1}}, \mathbf{d_{n+1}}} 
	\end{split}
\end{align}

These probabilities are the bedrock of all tasks. Namely, inference replaces the sums in eq. \eqref{eq:alpha} with ``max" operators plus backtracking (i.e., Viterbi algorithm \cite{viterbi1967error} or maximum a posteriori (MAP) estimate). Likewise, LL calculation reduces to $p(\mathbf{y};\text{\boldmath$\theta$\unboldmath}) = \sum_{\mathbf{z_{N}}} \sum_{\mathbf{d_{N}}} \alpha_{\mathbf{z_N}, \mathbf{d_N}}$. Next, by Bayes theorem and d--separation:

\begin{table*}[h!]
	\centering
	\begin{tabular}{|c |c|}
		\hline
		\parbox[][1.15cm][t]{6.5cm}{\begin{equation}
				\eta(\mathbf{z_{n}}, \mathbf{d_{n}})  = \mathunderline{yellow}{p(\mathbf{z_{n}}, \mathbf{d_{n}}|\mathbf{y})} = 
				\frac{\alpha_{\mathbf{z_n}, \mathbf{d_n}} \beta_{\mathbf{z_n}, \mathbf{d_n}}}{p(\mathbf{y};\text{\boldmath$\theta$\unboldmath}))} 
		\end{equation}}&
		\parbox[][1.15cm][t]{5.5cm}{\begin{equation}
				\gamma(\mathbf{z_{n}}) = \mathunderline{cyan}{p(\mathbf{z_{n}}| \mathbf{y})} =\sum_{\mathbf{d_{n}}}\eta(\mathbf{z_{n}}, \mathbf{d_{n}})
		\end{equation}}\\
		\hline
		\multicolumn{2}{|c|}{
			\parbox[][1.8cm][t]{12cm}{\begin{align}
					&\xi(z_{n,k}, \mathbf{z_{n-1}}, \mathbf{d_{n}}) = \mathunderline{green}{p(z_{n,k}, \mathbf{z_{n-1}}, \mathbf{d_{n}}, d_{n-1,1}|\mathbf{y}; \{\text{\boldmath$\lambda$\unboldmath}\}_{k=1}^K, A)}  = \\
					&\frac{\alpha_{\mathbf{z_{n-1}},d_{n-1,1}} p(y_n|\mathbf{z_n}) p(\mathbf{d_n};\text{\boldmath$\lambda$\unboldmath}_k) p(\mathbf{z_n}|\mathbf{z_{n-1}};A) \beta_{\mathbf{z_n}, \mathbf{d_n}} } {p(\mathbf{y};\text{\boldmath$\theta$\unboldmath})} \nonumber
		\end{align}}}\\
		\hline
	\end{tabular}
\end{table*}
where $p(y_n|\mathbf{z_n})$ is given by a conditional generalized t distribution. By construction, the entries of $\mathbf{z_{n}}$ and $\mathbf{d_{n}}$ are binary; hence, their expectations  are the probabilities of taking the value 1 (same color coding as eq. \eqref{eq:loglike}). 

For the M--step, we plug in the previous expectations into eq. \eqref{eq:loglike} and optimize each $\text{\boldmath$\theta$\unboldmath}$ component. The optimal parameters for the discrete hidden variables are simple global maximizers due to convexity  (i.e., partial derivatives equal to 0):
\begin{table*}[!]
	\centering
	\begin{tabular}{|c |c|}
		\hline
		\parbox{4cm}{\begin{equation}
				\pi_k^{(t)} \propto \gamma(z_{1,k})
			\end{equation}}&
		\parbox[][1.05cm][t]{8cm}{\begin{equation}
				A_{j,k}^{(t)} \propto \sum_{n=2}^{N} \sum_{\mathbf{d_{n}}} \xi(z_{n,k}, z_{n-1,j} ,\mathbf{d_{n}})
		\end{equation}}\\
		\hline
		\multicolumn{2}{|c|}{
			\parbox[][1.05cm][t]{12cm}{\begin{equation}
					\text{\boldmath$\lambda$\unboldmath}_{k,i}^{(t)} \propto \eta(z_{1,k},d_{1,i}) + \sum\limits_{n=2}^{N} \sum\limits_{\mathbf{z_{n-1}}} \xi(z_{n,k}, \mathbf{z_{n-1}}, d_{n,i})
				\end{equation}}}\\
		\hline
	\end{tabular}
\end{table*}

where $(t)$ denotes estimates for the $t$--th GEM iteration. $\gamma(z_{n,k})=\mathbb{E}\{z_{n,k}\}$ (analogous for $\eta$ and $\xi$). \boldmath$\lambda$\unboldmath$_{k,i}^{(t)}$ is the probability of duration $i$, regime $k$. The observation parameters are updated with local optima obtained via stochastic gradient descent and backpropagation applied to the last line of eq. \eqref{eq:loglike}. 

\section{Results}
\label{sec:results}

The DREAMS Sleep Spindles dataset \cite{stephanie_devuyst_2005_2650142} was used to validate the proposed model. 30--minute--long, single--channel (CZ-A1 or C3-A1) recordings from 8 subjects with corresponding sleep spindles expert scores (2 experts) are available. All traces were resampled to 50 Hz and z--scored prior to feeding them to the model. No bandpass filtering, artifact rejection, nor windowing were implemented. The data was then partitioned into 8 folds of training and test sets (i.e., 7 subjects for training, 1 for testing). The performance measure is the by--sample Matthews Correlation Coefficient (MCC) between MAP output and expert scores (on test set). We detail several model variants due to the modularity of the framework: 
\begin{enumerate}[label=\alph*)]
	\item Supervised setting (S, E1, E2) where the expert labels (either experts union, expert 1, or expert 2, resp.) are used to fit the model (i.e., no EM needed).
	\item Unsupervised setting (U) where no labels are used (i.e, EM required).
	\item Hidden Markov (HM) or hidden semi--Markov (HSM) latent state dynamics (i.e., setting $D=1$ and allowing self--transitions reduces the model to HM).
	 \item ``Linear + 2 Non--Linear" or ``3 Non--Linear " depending whether the location models, $\{F_{\text{\boldmath$\theta$\unboldmath}_{\mu, k}}(y_{n-p}^{n-1})\}_{k=1}^K$, are linear (Fig. \ref{fig:DNDBN}) or non--linear, respectively.
\end{enumerate}

We set $K=2$, $p=5$, and $D=1$ (HM) or $D=50\times15$ (HSM). We allowed self--transitions for non--spindles to accommodate long spindle--free intervals. For the deep models, we used stochastic gradient descent (batch size=32), early--stopping (75/25 split, patience=3, max epochs=10), and Adam \cite{kingma2014adam} ($\alpha=0.001$, $\beta_1=0.9$, $\beta_2=0.999$) on Python's tensorflow probability \cite{dillon2017tensorflow}. In each fold, a subject was held--out for validation and the best model out of 3 was kept for testing (according to MCC or LL for supervised or unsupervised cases, respectively).

Table \ref{table:Supervised} summarizes MAP results (without any post--processing) compared to three previous efforts \cite{wendt2012validation,martin2013topography,parekh2014sleep}.
The supervised--HSM setting has the highest average MCC, while the unsupervised setting outperforms two of the state--of--the--art methods. It is worth noting that the authors in \cite{parekh2014sleep} use window--based spectral methods to suppress artifacts in a pre--processing stage and all other methods in Table \ref{table:Supervised} bandpass the signal. Therefore, DNDBN not only performs feature extraction and detection, but it also handles outliers in a principled manner. All DNDBN types of Table \ref{table:Supervised} are ``Linear + 2 Non--Linear". Incidentally, the DNDBN(S, HSM) ``3 Non--Linear" variant achieves the same MCC of 0.459. As a side note, the approach in \cite{penny1999dynamic} achieves an average MCC of 0.071.

For  the unsupervised setting, we set sensible initial conditions: \begin{enumerate*}[label=\roman*)]
	\item $\text{\boldmath$\pi$\unboldmath} = [1, 0]^\top$, 
	\item $A = \bigl( \begin{smallmatrix}0.5 & 0.5\\ 1 & 0\end{smallmatrix}\bigr)$ (no self--transitions for sleep spindles),
	\item uniform durations for non--spindles and $\mathcal{N}(1 \text{ s.}, 0.15 \text{ s.})$ for spindles, and
	\item initial posterior marginals, $\gamma(\mathbf{z_{n}})$, equal to the average of the labels from 8 different detectors implemented in the package \textit{wonambi} (\url{https://github.com/ wonambi-python/wonambi}): six state--of--the--art \cite{ferrarelli2007reduced,molle2011fast,nir2011regional,leclercq2011fmri,wamsley2012reduced,martin2013topography} and two native to \textit{wonambi}. We refer to this mixture of experts type of detector as MixDetectors.
\end{enumerate*}

\begin{figure}[tb]
	\centering
	\includegraphics[width=0.99\linewidth]{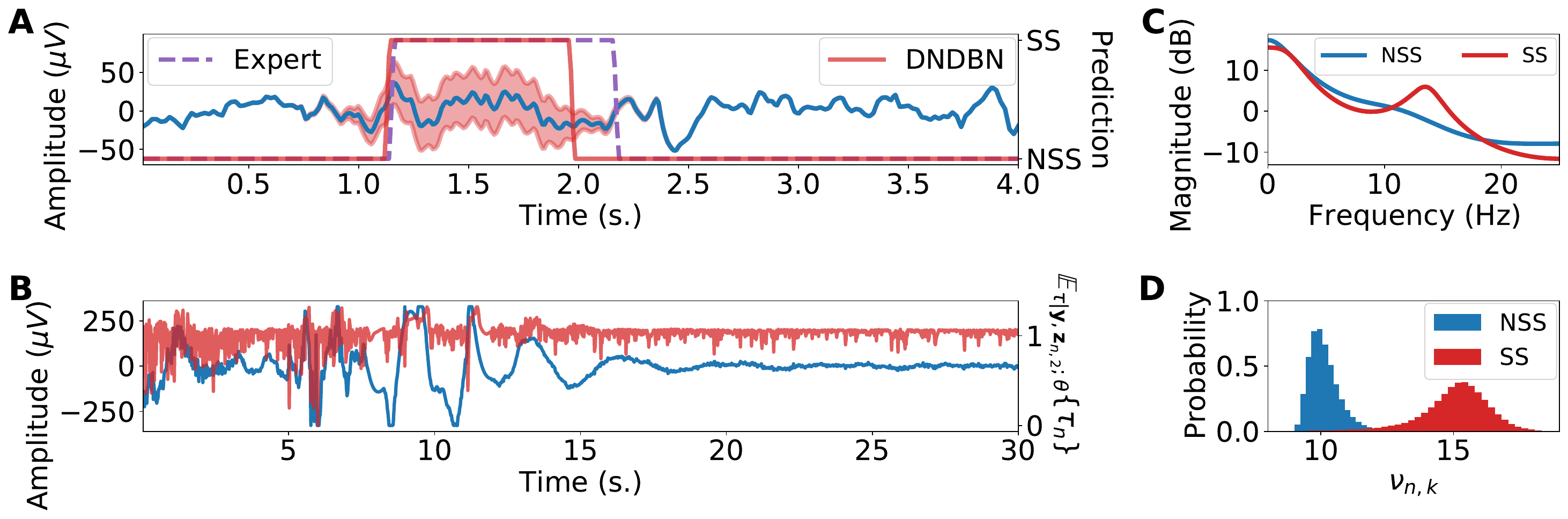}
	\caption{\textbf{A}. Example of MAP (red) and posterior marginal detections (shaded red, larger shaded areas imply larger probabilities of a sleep spindle). \textbf{B}. Example of expected value of $\tau_n$ under the posterior for EEG with and without artifacts. \textbf{C}. Log frequency response of learned location coefficients, ${\text{\boldmath$\theta$\unboldmath}_{\mu, k}}$. \textbf{D}. Histogram of learned degrees of freedom, $\nu_{n,k}$. SS: sleep spindle, NSS: non--sleep spindle.}
	\label{fig:Fig2}
\end{figure}

Fig. \ref{fig:Fig2} validates the model while illustrating its advantages over classic detectors. DNDBN is able to not only provide binary labels (MAP estimates), but it also quantifies the uncertainty via the posterior marginal, $\gamma(\mathbf{z_{n}})$. The model also provides a proxy for robustness via the expected value of the posterior $\tau_n$. In particular, by conjugacy, $\tau_n$ remains Gamma with updated means dependent of the observation parameters \cite{peel2000robust}. The first 15 seconds of Fig.  \ref{fig:Fig2}B show EEG plagued with artifacts, while the rest is clean EEG. Likewise, the posterior $\tau_n$ is able to track such changes while remaining bounded. Fig. \ref{fig:Fig2}C  shows the clear spectral differences between modes: the non--spindle regime depicts the well--known $1/f$ spectral distribution \cite{freeman2012imaging} while its counterpart has clear oscillatory activity in the sigma band. Lastly, the distributions of  $\{\nu_{n,k}\}_{k=1}^2$ confirm the expert opinion that most artifacts appeared during non--spindle epochs (i.e, smaller values imply longer tails of the generalized t distribution). These  results highlight the fact that detection is only one of the many attributes of the model.
\begin{table*}[tb]
	\centering
	\caption{MCC metrics (DREAMS Spindles data). Best results marked in bold.}
	\label{table:Supervised}
	\begin{tabular}{|c||c|c|c|c|c|c|c|c||c|}
		\hline
		\textbf{Algorithm} & \multicolumn{9}{c|}{\textbf{Subject}} \\
		\cline{2-10}
		& \textbf{S1} & \textbf{S2} &  \textbf{S3} & \textbf{S4} & \textbf{S5} & \textbf{S6} & \textbf{S7} & \textbf{S8} & \textbf{Mean}\\
		\hline
		\textbf{Wendt et al.} & 0.491 & 0.407 & 0.390 & 0.207 & 0.497 & 0.543 & 0.131 & 0.230 & 0.362 \\
		\hline
		\textbf{Martin et al.} & 0.437 & 0.548 & 0.407 & 0.212 & 0.514 & 0.586 & 0.229 & 0.259 & 0.399 \\
		\hline
		\textbf{Parekh et al.} & 0.501 & 0.575 & \textbf{0.540} & 0.245 & \textbf{0.574} & \textbf{0.624} & 0.253 & \textbf{0.269} &  0.447\\
		\hline
		\textbf{DNDBN(U, HSM)} & 0.565 & 0.483 & 0.450 & 0.265 & 0.549 & 0.610 & 0.199 & 0.233 & 0.419 \\
		\hline
		\textbf{DNDBN(S, HSM)} & \textbf{0.588} & \textbf{0.604} & 0.484 & \textbf{0.295} & 0.572 & 0.616 & \textbf{0.259} & 0.254 & \textbf{0.459} \\
		\hline
	\end{tabular}
\end{table*}

Next, different local solutions can be compared also. Table \ref{table:ComparisonModels} details predictive NLL and shows that the parameters set by expert 2 have stronger predictive power and follow the EEG dynamics closer (under the model in question). Also unsurprisingly, the ``3 Non--Linear" models outperform the ``Linear + 2 Non--Linear" counterparts. However, if we are interested in detection only, ``Linear + 2 Non--Linear" might suffice, as previously confirmed by the average MCC.

\begin{figure}[tb]
	\centering
	\includegraphics[width=0.99\linewidth]{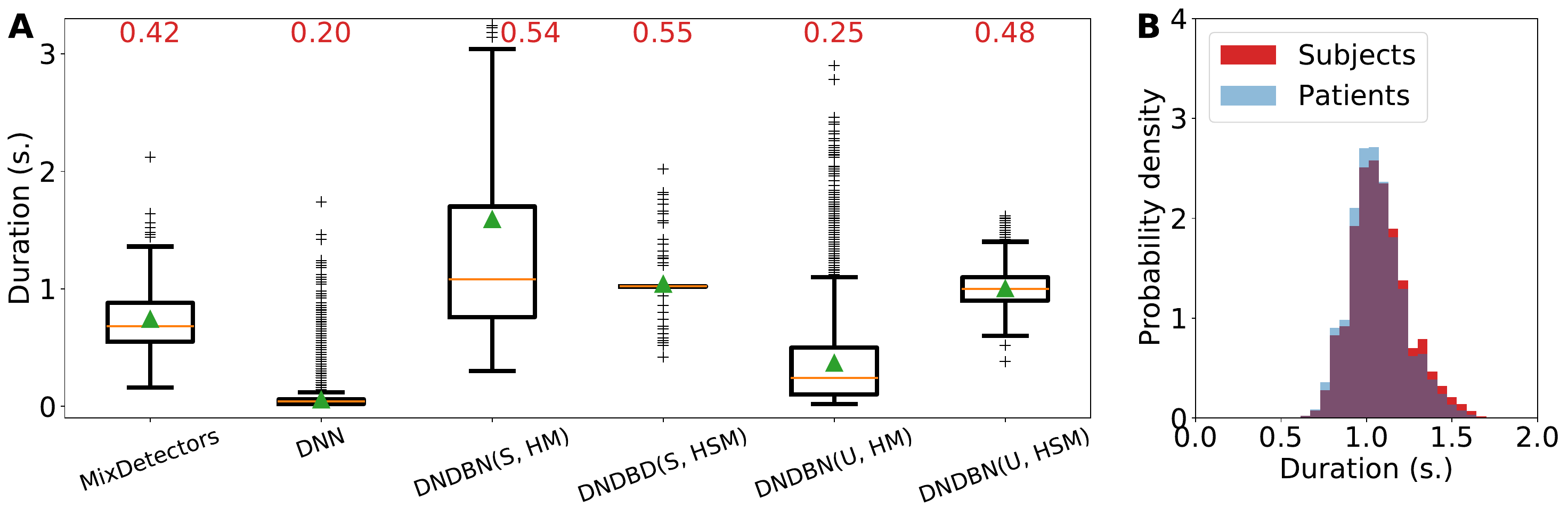}
	\caption{Distribution of sleep spindles durations. \textbf{A}. DREAMS Sleep Spindles dataset. Mean values marked with triangles. Areas under the precision--recall curves in red. Out--of--sample results with the exception of MixDetectors case. \textbf{B}. DREAMS Subjects and Patients datasets (stage N2 events only)}
	\label{fig:Fig3}
\end{figure}

Fig. \ref{fig:Fig3}A compares several models in terms of sleep spindles durations and area under the precision--recall curve (AUC) computed based on posterior marginals, $\gamma(\mathbf{z_{n}})$. An additional fully supervised deep learning detector was implemented as well (DNN) for benchmarking (6 128 fc layers + ReLU). Fig. \ref{fig:Fig3}A illustrates the interplay between interpretability, performance, and complexity. For instance, DNN performs poorly in both AUC and durations (no underlying probabilistic model). MixDetectors performs well, as expected. The HM variants seem to miss the mark when it comes to durations (see mean values). On the other hand, the HSM counterparts seem appropriate for both supervised and unsupervised cases, which suggests that a hidden semi--Markov state is indeed principled. 
\begin{table}[tb]
	\centering
	\caption{Average predictive negative log--likelihood (NLL). HSM variants.}
	\label{table:ComparisonModels}
	\begin{tabular}{|c||C{0.9cm}|C{0.9cm}|C{0.9cm}|C{0.9cm}||C{0.9cm}|C{0.9cm}|C{0.9cm}|C{0.9cm}|}
		\hline
		\textbf{Observation Model} & \multicolumn{4}{c||}{\textbf{Linear + 2 Non--Linear}} & \multicolumn{4}{c|}{\textbf{3 Non--Linear}} \\
		\hline
		\textbf{Type} & \textbf{S} & \textbf{E1} & \textbf{E2} & \textbf{U} & \textbf{S} & \textbf{E1} & \textbf{E2} & \textbf{U} \\  
		\hline
		\textbf{NLL}& 39876 & 40151 & 35312 & 39751 & 38553  & 38754  & 34192  & 38861 \\
		\hline
	\end{tabular}
\end{table}


The DREAMS Subjects and Patients datasets \cite{stephanie_devuyst_2005_2650142} are used for our final validation. The datasets consist of 20 and 27 whole--night recordings ($\approx 8-9$ hours), respectively. We train the model (DNDBN(U, HSM), ``Linear + 2 Non--Linear") on the DREAMS Sleep Spindles dataset and perform inference on both Subjects and Patients datasets (one central channel downsampled to 50 Hz). Fig \ref{fig:Fig3}B shows the duration distributions, which resemble previous  large--scale studies \cite{warby2014sleep}. In addition, Table \ref{table:SleepSpindlesRate} summarizes the sleep spindle rate (by--sample basis) for each sleep stage (scoring provided in the datasets). As expected, it is evident that stages N2 have relatively larger sleep spindles densities. Also, there are significant differences between datasets, which might be a result of the cohorts (i.e. Subjects: healthy individuals, Patients: individuals with various pathologies).

\section{Conclusion}
\label{sec:conclusion}
Probabilistic models over sequences are principled frameworks for robust detection and parametrization of  sleep spindles. Future work will go beyond the sigma band to characterize and detect high--frequency oscillations in other structures where ground truth is unavailable, e.g. hippocampal gamma rhythms.

\begin{table}[tb]
	\centering
	\caption{Average percentage of putative detected sleep spindles.}
	\label{table:SleepSpindlesRate}
	\begin{tabular}{|C{2cm}||C{1.8cm}|C{1.8cm}|C{1.8cm}|C{1.8cm}|C{1.8cm}|}
		\hline
		\textbf{Dataset} & \multicolumn{5}{c|}{\textbf{Sleep Stage}} \\
		\cline{2-6}
		& \textbf{wake} & \textbf{N1} & \textbf{N2} & \textbf{N3} & \textbf{REM} \\
		\hline
		\textbf{Subjects}& 4.26 & 7.01 & 13.54 & 8.65 & 6.23\\
		\hline
		\textbf{Patients}& 3.54 & 7.92 & 9.92 & 3.64 & 6.32\\
		\hline
	\end{tabular}
\end{table}


\bibliographystyle{splncs03}
\bibliography{bibfile}

\end{document}